\documentclass[letterpaper, 10 pt, conference]{ieeeconf}  

\IEEEoverridecommandlockouts                              
\usepackage{amsmath} 
\usepackage{amssymb} 
\usepackage{MnSymbol,wasysym} 
\usepackage{bbm} 

\usepackage{graphicx} 
\usepackage{epstopdf} 
\usepackage{tikz,pgfplots} 
\usepackage{pgfplotstable}
\usepgfplotslibrary{fillbetween}
\pgfplotsset{compat=1.18} 
\usepackage[caption=false]{subfig}
\usepackage{threeparttable}

\usepackage{multirow} 
\usepackage{multicol} 
\usepackage{booktabs} 
\usepackage{makecell} 

\usepackage{algorithm}  
\usepackage{algorithmic} 
\usepackage{mathrsfs}

\usepackage{textcomp}  
\usepackage{moresize} 
\usepackage{fontawesome} 
\usepackage[dvipsnames]{xcolor} 
\usepackage{float}  
\usepackage{calc} 

\usepackage{comment}  
\usepackage{acronym} 
\usepackage{cite}  
\usepackage{xparse}  

\usetikzlibrary{matrix}
\usetikzlibrary{positioning,chains}
\usetikzlibrary{shapes.geometric}
\usetikzlibrary{calc}
\usetikzlibrary{trees}
\usetikzlibrary{decorations.pathreplacing,calc,tikzmark, automata, arrows}
\usetikzlibrary{backgrounds, fit}
\usetikzlibrary{positioning,arrows,petri,shapes.misc}
\usetikzlibrary{arrows.meta, bending, calc, fit, shapes, patterns}

\definecolorset{RGB}{PLOT}{}{%
	Brown, 153, 79, 0;%
	Black,   0,   0,   0;%
	Blue, 86, 180, 233;%
	DarkBlue, 0, 101, 189;%
	Orange, 230, 159, 0;%
	DarkOrange, 213, 94, 0;%
	Purple, 93, 58, 155;%
	Pink, 220, 38, 127;%
	Red, 227, 27, 35;%
	Yellow, 255, 195, 37;%
	BluishGreen, 0, 158,  115;%
	ReddishPurple, 204, 121, 167;%
	Olive, 162, 173, 0;%
	Gray, 153, 153, 153;%
	Cyan, 64, 176, 166;%
	DarkPurple, 136, 34, 85 %
}

\pgfplotsset{
	compat=1.16,
	legend image code/.code={ 
		\draw[mark repeat=2,mark phase=2]
		plot coordinates {
			(0cm,0cm)
			(0.05cm,0cm)        
			(0.2cm,0cm)         
		};%
	}
}

\pgfplotsset{hide scale/.style={
		/pgfplots/xtick scale label code/.code={},
		/pgfplots/ytick scale label code/.code={}}}

\providecommand{\filepath}{.}
\newcommand{\figpath}{\filepath/data/}

\newcommand{\plotmeansolid}[3]{
	\addplot [
	mark=None,
	line width=1pt,
	color=#1,
	solid,
	] 
	table[
	x = Time,
	y = #3
	] 
	{\data};
	\addlegendentry{#2}
}

\newcommand{\plotmeansolidD}[3]{
	\addplot [
	mark=None,
	line width=1pt,
	color=#1,
	solid,
	] 
	table[
	x = ts,
	y = #3
	] 
	{\data};
	\addlegendentry{#2}
}

\newcommand{\plotmeandashedD}[3]{
	\addplot [
	mark=None,
	line width=1pt,
	color=#1,
	dashed,
	] 
	table[
	x = ts,
	y = #3
	] 
	{\data};
	\addlegendentry{#2}
}

\newcommand{\plotstdD}[3]{
	\addplot [
	mark=None,
	forget plot,
	line width=0pt,
	opacity=0,
	name path=A,
	] 
	table[
	x = ts, 
	y expr=(\thisrow{#2}) + (#3))
	] 
	{\data};
	
	\addplot [
	mark=None,
	forget plot,
	line width=0pt,
	opacity=0,
	name path=B,
	] 
	table[
	x = ts, 
	y expr=(\thisrow{#2}) - (#3))
	] 
	{\data};
	
	\addplot [
	forget plot,
	color=#1,
	fill opacity=0.2,
	] 
	fill between [
	of=A and B];
}

\newcommand{\plotmeansolidClinear}[2]{
	\addplot [
	mark=None,
	line width=1pt,
	color=#1,
	solid,
	] 
	table[
	x = Time,
	y = Mean
	] 
	{\data};
	\addlegendentry{#2}
}

\newcommand{\plotmeandashed}[3]{
	\addplot [
	mark=None,
	line width=1pt,
	color=#1,
	dashed,
	] 
	table[
	x = Time,
	y = #3
	] 
	{\data};
	\addlegendentry{#2}
}

\newcommand{\plotstdypred}[1]{
	\addplot[mark=None,
	forget plot,
	line width=0pt,
	opacity=0,
	name path=A] 
    coordinates {(-0.1,0.8)(5,0.8)};
	
	\addplot [
	mark=None,
	forget plot,
	line width=0pt,
	opacity=0,
	name path=B,
	] 
	coordinates {(-0.1,1)(5,1)};
	
	\addplot [
	forget plot,
	color=#1,
	fill opacity=0.2,
	] 
	fill between [
	of=A and B];

    	\addplot[mark=None,
	forget plot,
	line width=0pt,
	opacity=0,
	name path=C] 
    coordinates {(5, 0.95)(20, 0.95)};
	
	\addplot [
	mark=None,
	forget plot,
	line width=0pt,
	opacity=0,
	name path=D,
	] 
	coordinates {(5,0.85)(20, 0.85)};
	
	\addplot [
	forget plot,
	color=#1,
	fill opacity=0.2,
	] 
	fill between [
	of=C and D];
}

\newcommand{\plotclassSS}[3]{
	\begin{tikzpicture}
		\tikzstyle{every node}=[font=\scriptsize]
		\begin{axis}[hide scale, width=#2,
			height=#3,
			every tick label/.append style={font=\scriptsize},
			label style={font=\scriptsize},
			ymin=-0.05, ymax=1.05,
			xmin=0, xmax=18,
			xtick distance = 5,
			ytick={0.0,0.2,0.4,0.6,0.8,1.0},
			x tick label style={
				/pgf/number format/.cd,
				fixed,
				fixed zerofill,
				precision=0,
				/tikz/.cd
			},
			ylabel={concentration (a.u.)}, 
			xlabel={time (a.u.)},
			y label style={yshift=-.3em},
			x label style={yshift=.5em},
			legend style={at={(0.99, 0.01)},anchor=south east,font=\normalfont, nodes={scale=0.7, transform shape}},
			legend cell align={left}
			]
			{#1}
         \addplot[thick, dashed, samples=50, smooth,domain=0:6,gray] coordinates {(5,-0.1)(5,1)};
		\end{axis}
	\end{tikzpicture}
}

\newcommand{\plotclassD}[3]{
	\begin{tikzpicture}
		\tikzstyle{every node}=[font=\scriptsize]
		\begin{axis}[hide scale, width=#2,
			height=#3,
			every tick label/.append style={font=\scriptsize},
			label style={font=\scriptsize},
			ymin=-0.05, ymax=1,
			xmin=0, xmax=40,
			xtick distance = 5,
			ytick={0.0,0.2,0.4,0.6,0.8,1.0},
			x tick label style={
				/pgf/number format/.cd,
				fixed,
				fixed zerofill,
				precision=0,
				/tikz/.cd
			},
			ylabel={concentration (a.u.)}, 
			xlabel={time (hours)},
			y label style={yshift=-.3em},
			x label style={yshift=.5em},
			legend style={at={(0.99, 0.01)},anchor=south east,font=\normalfont, nodes={scale=0.7, transform shape}},
			legend cell align={left}
			]
			{#1}
		\end{axis}
	\end{tikzpicture}
}

\newcommand{\plotclassClinear}[5]{
	\begin{tikzpicture}
		\tikzstyle{every node}=[font=\scriptsize]
		\begin{axis}[hide scale, width=#2,
			height=#3,
			every tick label/.append style={font=\scriptsize},
			label style={font=\scriptsize},
			ymin=-20, ymax=#4,
			xmin=0, xmax=750,
			xtick distance = 250,
			ytick distance= #5,
			x tick label style={
				/pgf/number format/.cd,
				fixed,
				fixed zerofill,
				precision=0,
				/tikz/.cd
			},
			ylabel={concentration (a.u.)}, 
			xlabel={time (hours)},
			y label style={yshift=-.3em},
			x label style={yshift=.5em},
			legend style={at={(0.99, 0.01)},anchor=south east,font=\normalfont, nodes={scale=0.7, transform shape}},
			legend cell align={left}
			]
			{#1}
		\end{axis}
	\end{tikzpicture}
}

\newcommand{\PlotclassSS}[1]{
    \plotclassSS{
		\pgfplotstableread[col sep=comma]{\figpath #1.csv}{\data}
		\plotmeansolid{PLOTBlue}{$g (t)$}{y_pred}
        \plotstdypred{OliveGreen}
            \plotmeandashed{black}{$r(X_M,X_P)$}{y_true}
	}{4.6cm}{3.8cm}
}

\newcommand{\PlotclassD}[1]{
    \plotclassD{
		\pgfplotstableread[col sep=comma]{\figpath #1.csv}{\data}
		\plotmeansolidD{OliveGreen}{$g$}{y}
        \plotmeansolidD{PLOTBlue}{$x_1$}{x1}
        \plotmeansolidD{PLOTOrange}{$x_2$}{x2}
        \plotmeandashedD{black}{$r(X_M,X_P)$}{err}
        \plotstdD{OliveGreen}{err}{0.1}
        \legend{}
	}{4.6cm}{3.8cm}
}

\newcommand{\PlotclassClinear}[4]{
    \plotclassClinear{
		\pgfplotstableread[col sep=comma]{\figpath#1no-control_sp#2.csv}{\data}
		\plotmeansolidClinear{PLOTOrange}{no control}
        \pgfplotstableread[col sep=comma]{\figpath#1simple-control_sp#2.csv}{\data}
        \plotmeansolidClinear{PLOTBlue}{simple}
        \pgfplotstableread[col sep=comma]{\figpath#1nfc-control_sp#2.csv}{\data}
        \plotmeansolidClinear{OliveGreen}{BNN control}
        \legend{}
	}{4.6cm}{3.8cm}{#3}{#4}
}

\acrodef{rl}[RL]{reinforcement learning}
\acrodef{ppo}[PPO]{proximal policy optimization}
\acrodef{stl}[STL]{Signal Temporal Logic}
\acrodef{ode}[ODE]{ordinary differential equation}
\acrodef{pde}[PDE]{partial differential equation}
\acrodef{odes}[ODE]{ordinary differential equations}
\acrodef{pso}[PSO]{Particle Swarm Optimization}
\acrodef{gnn}[GNN]{Graph Neural Network}
\acrodef{nn}[ANN]{Artificial Neural Network}
\acrodef{bnn}[BNN]{Biomolecular Neural Network}
\acrodef{crn}[CRN]{Chemical Reaction Network}
\acrodef{rnn}[RNN]{Recursive Neural Network}
\acrodef{relu}[ReLU]{Rectified Linear Unit}
\acrodef{rna}[RNA]{ribonucleic acid}

\DeclareMathOperator*{\argmin}{arg\,min}
\def\always{\square}
\def\event{\lozenge}
\newcommand{\indexa}{1jl}
\newcommand{\indexb}{2jl}

\def\degradation{\beta}  
\def\sequestration{\gamma}       
\def\production{\nu}        

\definecolor{gene1}{RGB}{148,34,31}
\definecolor{gene2}{RGB}{62,114,116}
\definecolor{analogy}{RGB}{255,242,204}
\definecolor{analogy2}{RGB}{191,181,153}

\NewDocumentCommand{\SetName}{m}{\MakeUppercase{#1}}
\NewDocumentCommand{\SpeciesName}{mm}{\MakeUppercase{#1}_{#2}}

\NewDocumentCommand{\steadystate}{m}{\bar{#1}}

\NewDocumentCommand{\timeseries}{m}{\mathbf{#1}} 

\title{\LARGE \bf
STL-based Optimization of Biomolecular Neural Networks \\ for Regression and Control}

\author{Eric Palanques-Tost$^1$, Hanna Krasowski$^2$, Murat Arcak$^2$, Ron Weiss$^3$, Calin Belta$^4$
\thanks{$^1$E. Palanques-Tost is with Boston University, Boston, MA, USA {\tt\small ericpt@bu.edu}}
\thanks{$^2$H. Krasowski and M. Arcak are with University of California, Berkeley, CA, USA {\tt\small krasowski,arcak@berkeley.edu}}
\thanks{$^3$R. Weiss is with the Massachusetts Institute of Technology, Cambridge, MA, USA {\tt\small rweiss@mit.edu}}
\thanks{$^{4}$C. Belta is with the University of Maryland, College Park, MD, USA {\tt\small cbelta@umd.edu}}
}

\begin{document}

\maketitle

\begin{abstract}
Biomolecular Neural Networks (BNNs), artificial neural networks with biologically synthesizable architectures, achieve universal function approximation capabilities beyond simple biological circuits. However, training BNNs remains challenging due to the lack of target data.
To address this, we propose leveraging Signal Temporal Logic (STL) specifications to define training objectives for BNNs. We build on the quantitative semantics of STL, enabling gradient-based optimization of the BNN weights, and introduce a learning algorithm that enables BNNs to perform regression and control tasks in biological systems. Specifically, we investigate two regression problems in which we train BNNs to act as reporters of dysregulated states, and a feedback control problem in which we train the BNN in closed-loop with a chronic disease model, learning to reduce inflammation while avoiding adverse responses to external infections.  Our numerical experiments demonstrate that STL-based learning can solve the investigated regression and control tasks efficiently.
\end{abstract}

\section{Introduction}

Synthetic biology has revolutionized biotechnology by enabling the design of genetic networks that repurpose biochemical mechanisms to perform new functions \cite{voigt2020-SynBioApplications2020}. Foundational circuits, such as the genetic bistable toggle switch \cite{Collins2000-toggleSwitch} and the synthetic oscillator \cite{elowitz2000-oscillator}, have led to the development of genetic devices for a wide range of applications, including bioremediation \cite{wang2019-ecolisynbiobioremediation}, cancer therapies \cite{feins2019-cartcellssynbiocancer}, and feedback controllers \cite{aoki2019-universalIntegralFeedback}. However, these early circuits implement relatively simple functions, while many biological systems would benefit from more complex synthetic circuits.

Synthetic biology has also been used to design biological circuits that implement machine learning architectures, such as \acp{nn}\cite{qian2011-dnaBNN,fil2022-hebbianBNN}. These biological circuits are often studied \textit{in silico} using mathematical models of their reaction kinetics. For instance, studies \cite{moorman2019-dynamicalBNN, chen2021-dnazymeBNN, samaniego2021-cellularBNN} propose \ac{ode} models of reaction networks with designated input and output species, where the relation between steady-state concentrations of input and output species is determined by computations analogous to an \ac{nn} perceptron. These reaction networks can be arranged in feed-forward layers, forming a biomolecular counterpart to \acp{nn}, which we refer to as \acp{bnn}. However, training \acp{bnn} is often challenging due to the lack of well-defined target data. 

Other fields, such as robotics, address the lack of target data with heuristic rewards or cost functions. Recently, formal specifications such as \ac{stl} have been applied to reinforcement learning and control, enabling the design of interpretable policies under complex temporal constraints \cite{decastro2020-stlRL, haghighi2019-stlControl}. 
Quantitative \ac{stl} semantics have enabled gradient-based optimization of larger structures, such as \acp{nn}, with respect to \ac{stl} specifications \cite{leung2023-stlBackprop}. This is particularly helpful for systems where precise target trajectories are hard to define, such as biological systems. Yet, the use of \ac{stl} in biology is limited. Most temporal-logic-based methods focus on biological network synthesis \cite{goldfeder2019-stlSynthesis, bernot2010-stlGRN} using \ac{stl} indirectly for search or model checking. Recent work in \cite{krasowski2024-stldesign} uses direct gradient-based learning from \ac{stl} specifications to infer biological model structure and parameters, but the use of \ac{stl} for training \acp{bnn} remains mainly unexplored.

\begin{figure*}
     \centering
     \input{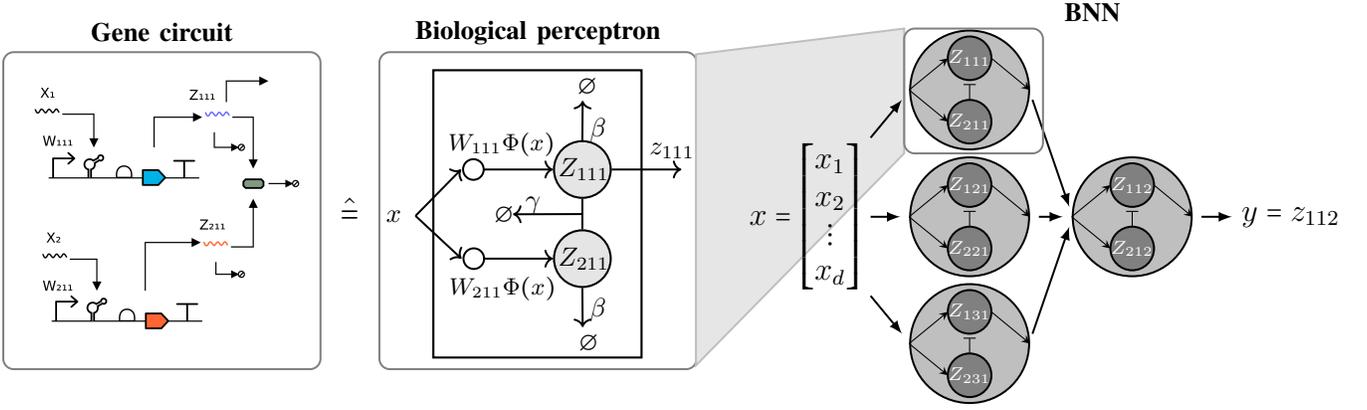}
     \caption{A \ac{bnn} is built from biological perceptrons implemented with synthetic genetic circuits. A possible genetic circuit implementing a \ac{bnn} perceptron is shown on the left.}
     \label{fig:bio_NN_pipeline}
\end{figure*}

In this paper, we present a training approach for \acp{bnn} using \ac{stl} specifications and we apply it in two tasks: (1) regression and (2) feedback control. In the regression task, the \ac{bnn} is trained to produce an output trajectory in response to a given input trajectory, such that the input and output satisfy an \ac{stl} specification. In the feedback control task, the \ac{bnn} is connected in closed-loop with a biological system, acting as a controller that drives the system toward a desired behavior specified by an \ac{stl} formula. We focus on \ac{bnn} training, instead of generic biological network synthesis, because \acp{bnn} provide a well-defined network structure composed of multiple instances of a single module (a biomolecular perceptron). These standardized architectures simplify the study and predictability of network behavior, as the interactions within and between modules follow a consistent, feed-forward pattern. 
The use of \ac{stl} formulae allows us to impose desired behaviors for the network without the need for numerical target data, which is often scarce in biological contexts. Our pipeline is end-to-end differentiable, enabling the optimization of the \ac{bnn} parameters with gradient-based techniques. The main contributions are:
\begin{itemize}
    \item We present a framework for gradient-based optimization of \ac{bnn} parameters to maximize the satisfaction of \ac{stl} specifications. 
    \item We formulate and solve two general problems using \acp{bnn}: regression and feedback control.
    \item We show three specific applications of our work, training \acp{bnn} for: (1) steady-state input regression to detect dysregulated states in a two-protein system, (2) dynamic input regression to identify transient dysregulation in a time-evolving two-protein system, and (3) feedback control of a chronic inflammation model with adaptive response to bacterial infection.
\end{itemize}

The remainder of this paper is organized as follows. In Sec.~\ref{sec:preliminaries}, we introduce the notation, provide an informal introduction to \ac{stl}, and define the \ac{bnn} used here. In Sec.~\ref{sec:problem-formulation}, we formulate the regression and control problems and in Sec.~\ref{sec:approach}  we describe the technical approach. In Sec.~\ref{sec:experiments}, we present our applications to static regression (Sec.~\ref{sec:static-clasification}), dynamical regression (Sec.~\ref{sec:dynamic-classification}) and closed-loop feedback control (Sec.~\ref{sec:results-control}). We discuss the results and give conclusions in Sec.~\ref{sec:discussion}.

\section{Preliminaries} 
\label{sec:preliminaries}

\subsection{Notation}
\label{sec:notation}

Indexed capital letters are used to denote species and corresponding lowercase letters to refer to their concentrations. Ordered sets of species concentrations are represented as vectors. For example, given an ordered set $S$ of $M$ species $S_i$, $i=1,\ldots,M$, their concentrations at time $t$ are represented as $s(t)=[s_1(t)\ldots s_M(t)]^T\in\mathbb{R}^M_{\ge 0}$, where $s_i(t)\ge 0$ is the concentration of $S_i$ and $[\cdot]^T$ denotes transposition. We assume the initial time is $0$ and the final time is $T>0$. Given a set of $N+1$ (discrete) time points $0=t_0<t_1\ldots<t_{N}=T$, 
with a slight abuse of notation, we use $s_i[k]:=s_i(t_k)$. 
A discrete time series of length $N+1$ for the concentrations of all species in the set is represented as ${\bf s}:=[s[0]\ldots s[N]]\in \mathbb{R}^{M\times (N+1)}$. Finally, the steady-state concentration of species $\SetName{s}_i$ is denoted by $\steadystate{s}_i$.

\subsection{Biomolecular Neural Network} 
\label{sec:bnn}

We consider the \ac{bnn} proposed in \cite{moorman2019-dynamicalBNN}, which implements \ac{nn}-like computations with a small number of species, using chemical reactions that are both common in nature and experimentally accessible. Notably, the network remains asymptotically stable at any depth. Each perceptron in this \ac{bnn} is composed of two biochemical species ($Z_1$ and $Z_2$), which inactivate each other at a rate $\gamma \in \mathbb{R}_{+}$. Species $Z_1$ and $Z_2$ degrade naturally at a rate $\degradation \in \mathbb{R}_{+}$. Given a set $X$ of $d$ input species with concentration $x\in\mathbb{R}_{+}^{d}$, the production rate of $Z_1$ and $Z_2$ depend on $x$ according to the equations  $\production_1(x)=W_1^T \cdot \Phi(x) + b_1$ and $\production_2(x)=W_2^T \cdot \Phi(x) + b_2$ respectively, with $W_1,W_2\in \mathbb{R}_{+}^d$, $b_1,b_2\in \mathbb{R}_{+}$ and $\Phi: \mathbb{R}^d\to \mathbb{R}^d$ being a function defined in the following paragraphs. The output of a perceptron is defined as the concentration of its species $\SpeciesName{z}{1}$. The reactions present in this perceptron can be implemented through a variety of mechanisms in biology, for instance, using a system of sense and anti-sense \ac{rna} that bind and form an inactive complex.
A schematic representation of the reactions in a perceptron, along with an example genetic circuit implementing it, is provided in Fig.~\ref{fig:bio_NN_pipeline}.

\acp{bnn} are constructed by stacking perceptrons in layers, where each perceptron receives the outputs from the previous layer. Imagine a \ac{bnn} with $L$ layers and $D_l$ perceptrons in layer $l$. We denote $\SetName{y}$ the input species of the full \ac{bnn}, and its output species $U$. Here, $x$ will be input of the first layer of the \ac{bnn}, and $y=\{z_{1,j,L} | j=1, \ldots, D_L\}$ will be the concentration of the output species in the last layer. Let us define the vector $h_l=[z_{1,1,l}, z_{1,2,l}, \ldots, z_{1,D_{l},l}]$ with the concentration of all the output species $Z_1$ in a layer $l$, with $y=h_L$. The dynamics of the perceptrons $j=1,\ldots,D_l$ in layer $l=1,\ldots,L$, containing species $\SpeciesName{z}{\indexa}$ and $\SpeciesName{z}{\indexb}$, are described by Eq.~\eqref{eqn:bnn-dynamics}, where we take $h_0=x$: 
\begin{equation} \label{eqn:bnn-dynamics}
    \begin{aligned}
        &\begin{aligned}
            \dot{z}_{\indexa} &= W^T_{\indexa} \cdot\Phi_l(h_{l-1}) + b_{\indexa}-\sequestration \,z_{\indexa}\,z_{\indexb} -\degradation \,z_{\indexa},
            \\
            \dot{z}_{\indexb} &= W^T_{\indexb}\cdot\Phi_l(h_{l-1}) + b_{\indexb}  -\sequestration \,z_{\indexa}\,z_{\indexb} -\degradation \,z_{\indexb}.
        \end{aligned}
    \end{aligned}
\end{equation}

The function $\Phi_l$ is the result of applying $\phi(\cdot)$ to every element of $h$, specifically: $\Phi_l(h_{l-1})=[\phi_l(z_{1,1,l-1}), \ldots, \phi_l(z_{1,D_{l-1},l-1})]$, with $\phi_1(x_i)=x_i$ in the first layer, and $\phi_l(z_i)=\frac{z_i}{k+z_i} \; \forall l\in\{2, \ldots, L\}$. The use of a different $\phi$ in the first layer is motivated by the different biological reactions required to process the external inputs. 

The dynamics of the full \ac{bnn} are defined by aggregating the dynamics of all its perceptrons for $j=1,\ldots, D_l$ and $l=1,\ldots,L$. The perceptrons in layers $l=1, \ldots, L$ on the \ac{bnn} have parameters $\theta_{j,l}=(W_{\indexa}, W_{\indexb}, b_{\indexa}, b_{\indexb})$, where $W_{\indexa}, W_{\indexb} \in \mathbb{R}_+^d$ and $b_{\indexa}, b_{\indexb} \in \mathbb{R}_+$, as well as parameters $\beta, \gamma\in \mathbb{R}_+$. Perceptrons in layers $l=2,\ldots,L$ also have the parameter $k\in \mathbb{R}_+$. Collectively, we denote the full parameter set of the \ac{bnn} $\Theta = \{\theta_{j,l}, j=1\ldots,D_l, l=1,\ldots,L\} \cup \{\gamma,k,\beta \}$. We use $\mathcal{B}_\Theta$ to refer to the dynamical system of a \ac{bnn} formed by Eq.~\eqref{eqn:bnn-dynamics} and parameters $\Theta$.

\subsection{Signal Temporal Logic}
\label{sec:stl}

We use \ac{stl} formulas to define temporal behaviors and logical dependencies among species in a biological system (see  \cite{maler2004-stlOriginal} for formal definitions of \ac{stl} syntax and semantics). Informally, \ac{stl} formulas consist of three ingredients: (1) predicates $\mu:=g(s)>0$, with $g:\mathbb{R}^M\rightarrow\mathbb{R}$; (2) Boolean operators, such as negation $\neg$, conjunction $\land$, and disjunction $\lor$; and (3) temporal operators, such as eventually $\event_{[k_1,k_2]}$ and always $\always_{[k_1,k_2]}$, where $k_1$, $k_2$ are two discrete time points with $k_1<k_2$. Given a predicate $\timeseries{\mu}$, $\event_{[k_1,k_2]}\timeseries{\mu}$ is true if $\timeseries{\mu}$ is satisfied for at least one time point $k\in[k_1,k_2]$, while $\always_{[k_1,k_2]}\timeseries{\mu}$ is true if $\timeseries{\mu}$ is satisfied for all $k\in[k_1,k_2]$. 

The semantics of \ac{stl} formulas is defined over time series ${\bf s}$ (see Sec.~\ref{sec:notation}). For example, formula $\event_{[2, 7]} s_1>0.5$ is satisfied over ${\bf s}$ if the concentration $s_1$ of species $S_1$ exceeds 0.5 at any time between 2 and 7; 
\ac{stl} also has quantitative semantics defined by the robustness function $\rho({\bf s}, \varphi)$, which measures how strongly ${\bf s}$ satisfies or violates $\varphi$; specifically, $\rho({\bf s}, \varphi) > 0$ iff ${\bf s}$ satisfies $\varphi$ \cite{donze2010-stlRobustness}.

\section{Problem Statement} \label{sec:problem-formulation}

Given a BNN with internal species $Z$ and dynamics governed by $\mathcal{B}_\Theta$ (Eq.~\eqref{eqn:bnn-dynamics}), we formulate two problems: (1) regression and (2) feedback control.

\subsection{Regression} \label{sec:problem-formulation-regression}

In the regression setting, the \ac{bnn} receives input species $X$ and produces output species $Y$. A desired input/output behavior is specified via an \ac{stl} formula $\varphi$, defined over the trajectories $\mathbf{x}$ and $\mathbf{y}$ of the input and output species, respectively. The objective is to find the parameters $\Theta$ of the \ac{bnn} such that the resulting \ac{bnn} output trajectory $\mathbf{y}$ satisfies $\varphi$ for the largest possible set of admissible input trajectories $\mathbf{x}$. For example, a \ac{bnn} may sense $X_1$ and $X_2$, and output a detectable biomarker $Y$. We may require $y$ to track $x_1+x_2$ over the first five time steps, $\varphi: \always_{[0,5]}|y-(x_1+x_2)|<0.1$, and optimize $\Theta$ so that $\varphi$ holds for as many $\mathbf{x}_1$ and $\mathbf{x}_2$ trajectories as possible.

\subsection{Feedback control} \label{sec:problem-formulation-control}

In feedback control, the \ac{bnn} is connected in a closed-loop configuration with a biological system composed of species $X$, whose dynamics are governed by $\dot{x}=f_p(x, u, t)$, where $p \in \mathcal{P}$ are the parameters of the \acp{ode}, and $u$ is the concentration of the \ac{bnn}'s output species. The \ac{bnn} receives as input a subset of species $Y$, where $y = g(x)$, and produces output species $U$ that act as control inputs. Given an \ac{stl} formula $\varphi$ defined over the system species trajectories $\mathbf{x}$, the objective is to find the parameters $\Theta$ of the \ac{bnn} such that the closed-loop trajectories satisfy $\varphi$ for the largest subset of initial conditions $x_0 \in \mathcal{X}_0$ and parameter values $p\in \mathcal{P}$, i.e. maximize the size of the subset of $X_0\times \mathcal{P}$ for which $\varphi$ is satisfied. For example, in a system with a damage marker $X_1$ and a repair factor $X_2$ that reduces $X_1$ at a rate $p$, a \ac{bnn} may be trained to control $x_1$ by increasing $x_2$ via its action $u$. We may impose $x_1$ to remain below 2.0 pmol, $\varphi: \always_{[0, \infty)} x_1 < 2.0$, and optimize $\Theta$ so that $\varphi$ is satisfied across as many initial concentrations $x_1(t_0), x_2(t_0)$ and repair rates $p$ as possible.

\section{STL-based optimization of BNNs} \label{sec:approach} 

We propose an optimization-based approach to solve the problems stated in Sec.~\ref{sec:problem-formulation}.

\subsection{Generation of a training and testing set} \label{sec:generate-diff-conditions}
For both regression and feedback control tasks, the continuous space of all possible inputs and parameters is intractable. Therefore, we construct a finite training set $\mathcal{X}$ by sampling from biologically plausible ranges. For regression, we generate $C$ input trajectories drawn from representative temporal patterns $\mathcal{X} = \{ \mathbf{x}^{c}, c=1, \ldots, C\}$ . For feedback control, we sample a set of initial conditions and plant parameters from biologically relevant values, $\mathcal{X} = \{(x^{c}(t_0), p^{c}), c=1, \ldots, C\}$. These ranges can be informed by prior knowledge of the system, and will be used to optimize and evaluate the \ac{bnn}.

\subsection{Integration of the dynamics} \label{sec:integration-of-dynamics}

Given a system as defined in Sec.~\ref{sec:problem-formulation}, whether for feedback control or regression, we simulate its behavior by numerically integrating its dynamics over a finite time horizon. Let $t = [t_0, \ldots, t_N]$ denote a vector of time points at which the system state is evaluated. In the regression problem, we compute the output time series $y[k]$ for all $k = 0, \ldots, N-1$ by integrating the \ac{bnn} dynamics over time.

To obtain the continuous time trajectories $x(t)$ required in the \ac{ode} solver, we apply linear interpolation to the discrete input trajectories. In the feedback control task, we obtain the state trajectories $x[k]$ for all $k = 0, \ldots, N-1$ by numerically integrating the closed-loop system, $f_p$ and $\mathcal{B}_{\Theta}$, over time with the initial conditions $x(t_0)$, $z(t_0)$.

\subsection{Optimization} \label{sec:approach-optimization}

Because solving the problem stated in Sec. \ref{sec:problem-formulation} over the full continuous space is infeasible, we instead minimize a loss function $\mathcal{L}$ on a discretized training set of size $C$:
\begin{equation} \label{eqn:loss-function}
\begin{aligned}
    &\Theta^* = \argmin_{\Theta} \mathcal{L}(\Theta), \\
    &\text{with: }\mathcal{L}(\Theta) = \sum_{c=1}^C \max \bigr(0, -\rho(\varphi, \timeseries{s}^c|\Theta)\bigl).
\end{aligned}
\end{equation}
In this equation, $\timeseries{s}^c$ are the trajectories over which the \ac{stl} formula $\varphi$ is defined (i.e. the \ac{bnn} input and output in regression and the system species in feedback control), which depend on the \ac{bnn} parameters $\Theta$. The loss function penalizes only trajectories violating $\varphi$ (i.e. $\rho < 0$), while those satisfying $\varphi$ (i.e. $\rho>0$) make no contribution, so they cannot counterbalance violations.

The optimization procedure in Alg.~\ref{alg:optimization} begins with a training set $\mathcal{X}$ containing $C$ different conditions (Sec. \ref{sec:generate-diff-conditions}), either input trajectories (in regression) or combinations of initial conditions and system parameters (in feedback control); an \ac{stl} formula $\varphi$ with the desired system behavior; and a vector of time points $t$ to evaluate the system (line 1). For each condition in $\mathcal{X}$, the system dynamics are integrated as in Sec. \ref{sec:integration-of-dynamics}  (line 3), the loss $\mathcal{L}$ is computed as in Eq.~\eqref{eqn:loss-function} (line 4), and the robustness~$\rho$ is evaluated as in~\cite[Eq.~(2)]{leung2023-stlBackprop}. We use the non-smooth robustness formulation to avoid smoothing errors, ensuring that the \ac{stl} specification is truly satisfied when $\rho \ge 0$. If $\mathcal{L}>0$ (lines 5–7), the \ac{bnn} parameters are updated via gradient descent using the AdaBelief optimizer~\cite{zhuang2020-adabelief} (line 9), which is well-suited for non-convex objectives such as ours.
Gradients are computed via automatic differentiation through both the \ac{ode} solver and the robustness function, avoiding inaccuracies that can arise from the adjoint sensitivity method~\cite{chen2018-neuralODE}. The parameters $\Theta$ are optimized in log-space $\tilde{\Theta} = \log(\Theta + 10^{-5})$ (line 8) and mapped back after each update as $\Theta=\exp(\tilde{\Theta})-10^{-5}$ (line 10), ensuring they remain non-negative, and keeping the system stable. Finally, we evaluate the \ac{bnn} performance on a separate test set to assess its generalization beyond the training data.

\begin{algorithm}[tb]
\caption{Gradient-based BNN learning with \ac{stl}}
\label{alg:optimization}
\begin{algorithmic}[1]
    \STATE \textbf{Initialize:} BNN $\mathcal{B}_{\Theta}$ with $L$ layers, $D_l$, $l=1, \ldots, L$ perceptrons per layer, parameters $\Theta$,  biological system with dynamics $f_p$, problem-specific condition set $\mathcal{X}$, hyperparameter $\lambda$, maximum iterations $I_{\mathrm{max}}$
    \FOR{$i = 1$ \textbf{to} $I_{\mathrm{max}}$}
        \STATE Obtain $\{(\timeseries{s}^c|{\Theta}), c=1,\ldots, C\}$ for all conditions $\mathcal{X}$.
        \STATE Compute loss: $\mathcal{L} = \sum_{c=1}^C \max(0, -\rho(\varphi, \mathbf{s}_{\Theta}^c))$
        \IF{$\mathcal{L} = 0$}
            \STATE \textbf{break}
        \ENDIF
        \STATE Project the weights to log space: $\tilde{\Theta} = \log(\Theta+10^{-5})$
        \STATE Update parameters: $\tilde{\Theta} \gets \mathtt{AdaBelief}_{\lambda}(\tilde{\Theta}, \mathcal{L})$
        \STATE Undo the weight projection: $\Theta = \exp(\tilde{\Theta})-10^{-5}$

    \ENDFOR
    \RETURN BNN $\mathcal{B}_{\Theta}$
\end{algorithmic}
\end{algorithm}

\section{Experiments} \label{sec:experiments}

We evaluate our \ac{stl}-based \ac{bnn} optimization framework on three examples. 
All experiments are implemented in JAX \cite{jax2018-github} with diffrax \cite{kidger2021-diffrax} for numerical integration. 
\ac{bnn} parameters are initialized using a Xavier/Glorot initialization \cite{glorot2010-xavierInit} truncated to 0 for negative values, and with small positive biases to improve convergence. We globally set $\sequestration=1000$, $\degradation=1.0$, and $k=0.8$, using the Kverno5 \ac{ode} solver \cite{kvaerno2004-sdirk}.

\subsection{Static input regression} \label{sec:static-clasification} 

In this example, we consider a biological system with two proteins: $p53$, which triggers apoptosis (cell death), and $Mdm2$, which inhibits apoptosis by targeting $p53$ for degradation. We denote their concentration as $x_P$ and $x_M$, respectively. To maintain cell health, $x_P \approx x_M$. Therefore, we train a \ac{bnn} with three perceptrons (two in the first layer, and one in the output layer) to report their dysregulation via a fluorescent protein output, $G$, governed by $\dot{g}=\alpha\cdot \frac{y}{k+y} - \delta g$, with $\alpha=5$, $k=0.8$ and $\delta=1$. This reaction can be seen as an additional post-processing layer of the \ac{bnn}. For $x_P$ and $x_M$ typically lying in $[0, 1]$ in arbitrary units (a.u), we want $\bar{g}$  to approximate the function $r(x_M,x_P) = \max(0, |x_M-x_P|-0.1)$. The error should be below $0.1$ before $t=5$, and eventually below $0.05$: $\varphi_1: (\event_{[0, 5]} \always_{[0, \infty]} |g - r(x_M, x_P)| < 0.1) \land (\event_{[0, \infty]} \always_{[0, \infty]} |g - r(x_M, x_P)| < 0.05)$.

The training data is obtained by discretizing the input space $\mathcal{X}_P=\mathcal{X}_M=\{0, 0.1, \ldots, 1.0\}$ to form a grid $\mathcal{X}=\mathcal{X}_P\times\mathcal{X}_P$ of $C=121$ input pairs. The test set is obtained by shifting the grid 0.05 units. For each $(x_P, x_M)$ in the grid, we simulate trajectories of $g(t|\Theta)$ at $t =[0, 1, ..., 20]$ with constant $x_P$ and $x_M$ (i.e. $x_P(t)=x_P$ and $x_M(t)=x_M$). Here, each trajectory $g(t|\Theta)$ generated corresponds to $\mathbf{s}^c|\Theta$ in Eq.~\eqref{eqn:loss-function}, and the objective is to minimize that loss. 

\begin{figure}[h]
    \centering
    \begin{minipage}[t]{0.48\linewidth}
        \centering
            \def\svgwidth{\linewidth}
            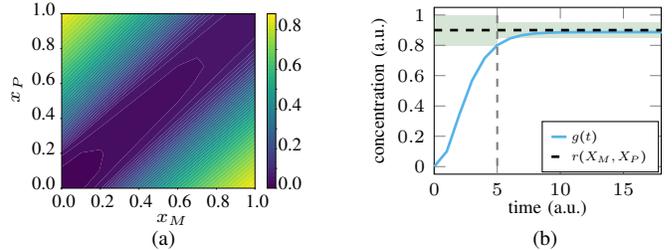 \vspace{0.3cm}
            {\footnotesize \hspace{-0.5cm} (a)}
    \end{minipage}%
    \hspace{0.01em}
    \begin{minipage}[t]{0.48\linewidth}
        \centering
            \raisebox{-1.0ex}{\PlotclassSS{ss_classification/traj}}
            \vspace{0.3cm}
            {\footnotesize \hspace{1.2cm}(b)}
    \end{minipage}
    \vspace{-0.3cm}
    \caption{Results of the static-input regression experiment. (a) Heatmap of the concentration of $g$ in steady state as a function of $X_M$ and $X_P$. (b) Evolution of $g$ over time for $X_M=0$ and $X_P=1$. The area where $\varphi$ is satisfied is shaded in green.
    }
    \label{fig:ss-classification-results}
\end{figure}

We train for a maximum of $3000$ iterations with an initial learning rate of $0.05$, halved every $1000$ iterations. Fig.~\ref{fig:ss-classification-results}(a) shows $\bar{g}$ as a function of $x_M$ and $x_P$, closely matching $r(x_M, x_P)$. Fig.~\ref{fig:ss-classification-results}(b) shows the transient behavior $g(t)$ for $x_M=0$ and $x_P=1$ satisfying the temporal constraints. To assess robustness, the optimization is repeated with $10$ different random seeds, $7$ of which converge (Tab.~\ref{tab:control-results-table}) and achieve $100\%$ satisfaction of $\varphi$ on the test set.  


\subsection{Dynamic input regression} \label{sec:dynamic-classification}

In this example, we extend the previous regression example to a dynamic setting where $x_M$ and $x_P$ vary over time. The goal is for $g$ to continuously approximate $r(x_M,x_P) = \max(0, |x_M-x_P|-0.1)$ within the next 9h, allowing it to ignore brief fluctuations during that window. The specification is $\varphi_2: \always_{[0, \infty]} \event_{[0, 9]} |y-r(x_P, x_M)|<0.1$.

We design a 2-layer \ac{bnn} with two perceptrons in the first layer, and one in the output layer. We create a training set with synthetic trajectories $x_P[k]$ and $x_M[k]$ for $k = 1, \ldots, 30$, generated with a two-state Markov chain: low and high. Continuous-value input trajectories are obtained via random walk with noise clipped to the current state: low $[0, 0.25]$ and high $[0.6, 1.0]$. Transitions between states are smoothed using linear interpolation over 3 steps preceding the change. We generate three sets of $50$ trajectory pairs $(x_P[k], x_M[k])$, each under the following conditions: (1) $x_P$ fixed in the low state, and $x_M$ switching states with probability 0.2, (2) like previous but reversed roles, (3) both $x_M$ and $x_P$ switching with probability 0.1. This results in a training set $\mathcal{X}$ with $C=150$ input trajectories. A test set of the same size is generated using a different random seed. For each input trajectory, the corresponding output trajectories $g(t|\Theta)$ are sampled at $t=[0, 0.5, 1, \ldots 40]$. We train for a maximum of $400$ iterations with a learning rate of $0.05$.

\begin{figure}[h]
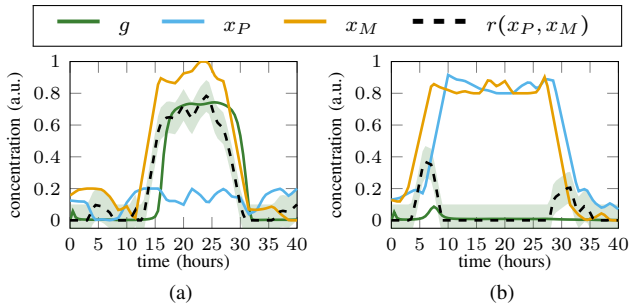

    \centering
\begin{footnotesize}
\hspace{0.7cm}\fbox{\begin{tabular}{l l l l }
    \textcolor{OliveGreen}{\textbf{\LARGE{---}}} \hspace{0.2em} $g$
    & \textcolor{PLOTBlue}{\textbf{\LARGE{---}}} \hspace{0.2em} $x_P$ 
    & \textcolor{PLOTOrange}{\textbf{\LARGE{---}}} \hspace{0.2em} $x_M$
    & \textcolor{black}{\textbf{\LARGE{-\,-\,-}}} \hspace{0.2em} $r(x_P, x_M)$\\ 
\end{tabular}
}

\end{footnotesize}
   \begin{minipage}[t]{0.48\linewidth}
        \centering
             \raisebox{-1.0ex}{\PlotclassD{dyn_classification/traj_100}}
            \vspace{0.3cm}
            {\footnotesize \hspace{1.2cm}(a)}
    \end{minipage}%
    \hspace{0.01em}
    \begin{minipage}[t]{0.48\linewidth}
        \centering
            \raisebox{-1.0ex}{\PlotclassD{dyn_classification/traj_6}}
            \vspace{0.3cm}
            {\footnotesize \hspace{1.2cm}(b)}
    \end{minipage}
    \vspace{-0.3cm}
  \caption{Time-series generated in the dynamic-input regression experiment. 
  The green shaded area represents the tolerated error $\epsilon=0.1$ at each time instance.}
  \label{fig:dyn-class-results} 
\end{figure}

The optimization is repeated with $10$ different seeds, of which $4$ achieve a training set satisfaction $\ge 90\%$ (Tab.~\ref{tab:control-results-table}). In these $4$ runs, the mean satisfaction in the test set is $94\%$. Fig.~\ref{fig:dyn-class-results} shows two representative trajectories. In Fig.~\ref{fig:dyn-class-results}(a), $g$ increases because $|x_P-x_M|$ remains high for $>9$ hours. In Fig.~\ref{fig:dyn-class-results}(b), short imbalances where $|x_P-x_M|>0.1$ do not raise $g$ significantly, since balance is quickly restored, showing that the \ac{bnn} can ignore short deviations.

\subsection{Closed-loop control} \label{sec:results-control}
\begin{figure*}
    \centering
    \begin{minipage}[t]{0.46\textwidth}
            \begin{tikzpicture}[auto,
  node distance = 7mm and 8mm,
  myarrow/.style={->,thick,shorten <=2pt,shorten >=2pt}]

  \tikzstyle{state}=[draw, rounded corners=1.2mm,
    inner ysep=1.0mm, inner xsep=1.0mm,
    minimum height=10mm, minimum width=1.4cm,
    align=center]

  \node[state] (bacteria) {\footnotesize Bacteria\\$X_B$};
  \node[state, right=12mm of bacteria] (pro) {\footnotesize Pro-Inflammatory\\$X_P$};
  \node[state, right=12mm of pro] (damage) {\footnotesize Damage\\$X_D$};

  \draw[myarrow] ([yshift=-1mm]bacteria.east) -- ([yshift=-1mm]pro.west);
  \draw[myarrow,-|] ([yshift=1mm]pro.west) -- ([yshift=1mm]bacteria.east);
  \draw[myarrow] ([yshift=-1mm]pro.east) -- ([yshift=-1mm]damage.west);
  \draw[myarrow] ([yshift=1mm]damage.west) -- ([yshift=1mm]pro.east);
  
  \node[state, below=9mm of pro] (anti) {\footnotesize Anti-Inflammatory\\$X_A$};
  \node[state, draw=OliveGreen, fill=OliveGreen!20, below=9mm of bacteria] (bnn)
       {\footnotesize BNN\\$\mathcal{B}_\Theta$};

  \draw[myarrow] ([xshift=1.5mm]pro.south) -- ([xshift=1.5mm]anti.north);
  \draw[myarrow,-|] ([xshift=-1.5mm]anti.north) -- ([xshift=-1.5mm]pro.south); 
  \draw[myarrow] (damage.south) |- (anti.east);
  \draw[myarrow] ([yshift=1.5mm]anti.east) -| ([xshift=-1.5mm]damage.south);

  \draw[myarrow] (bacteria.south) -- (bnn.north);
  \coordinate (lane) at ($(anti.south)+(0,-3mm)$); 
\draw[myarrow]
  ([xshift=1.5mm]damage.south) -- ([xshift=1.5mm]lane -| damage.south) -- (lane -| bnn.south) -- (bnn.south);

  \draw[myarrow] (bnn.east) -- (anti.west);

  \draw[myarrow] ($(bacteria.north)+(0,4mm)$) node[above] {\footnotesize Infection}
       -- (bacteria.north);
  \draw[myarrow] ($(damage.north)+(0,4mm)$) node[above] {\footnotesize Chronic damage}
       -- (damage.north);
\end{tikzpicture} 
    \end{minipage} 
    \begin{minipage}[t]{0.52\textwidth}
            \vspace{-3.9cm}
\begin{footnotesize}
\hspace{1.0cm}
\fbox{\begin{tabular}{l l l }
    \textcolor{OliveGreen}{\textbf{\LARGE{---}}} \hspace{0.2em} BNN
    & \textcolor{PLOTBlue}{\textbf{\LARGE{---}}} \hspace{0.2em} BNN (B-unaware) 
    & \textcolor{PLOTOrange}{\textbf{\LARGE{---}}} \hspace{0.2em} no control \\ 
\end{tabular}
} 
\end{footnotesize}
\vspace{0.25cm}

\begin{minipage}{0.48\linewidth}
   \centering
   \textbf{\small Bacteria ($\mathbf{X}_B$)}\\[0.1cm]
   \raisebox{-1.0ex}{\PlotclassClinear{control/}{0}{650}{200}}
\end{minipage}%
\hspace{0.02\linewidth}%
\begin{minipage}{0.48\linewidth}
   \centering
   \textbf{\small Damage ($\mathbf{X}_D$)}\\[0.1cm]
   \raisebox{-1.0ex}{\PlotclassClinear{control/}{3}{800}{200}}
\end{minipage}

    \end{minipage}
    
     \caption{Results of the optimization of the controller. Left: Diagram of the interactions in the biological model with triangle arrowheads for activation and T-bar arrows for inhibition. Right: Trajectories generated with $X_{B0}=407$ and $p=15$ in three scenarios: untreated (orange), treated with a bacteria-unaware \ac{bnn} controller (blue), and treated with a bacteria-aware \ac{bnn} controller (green).}
     \label{fig:control-results}
 \end{figure*}
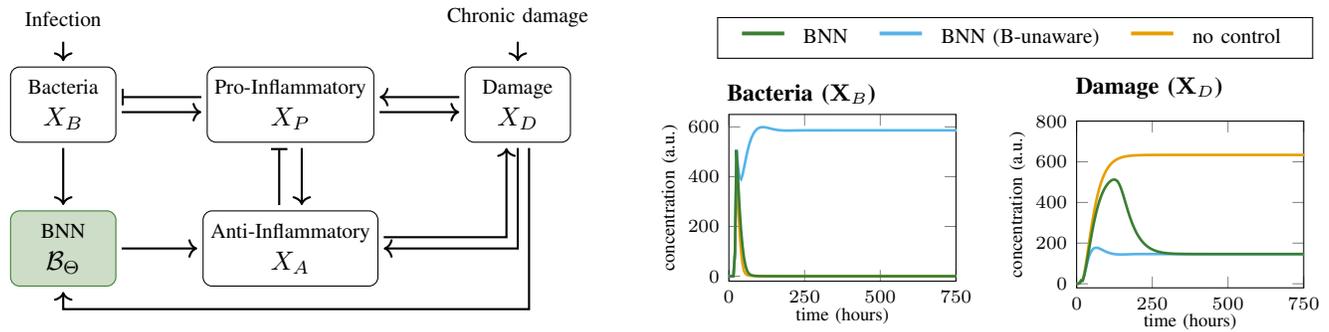

\begin{table}[]
    \centering
    \begin{threeparttable}[]
        \caption{Results from the training process}
        \label{tab:control-results-table}
        \begin{tabular}{c c c c}
            \toprule
            \textbf{Problem} &
            \begin{tabular}[c]{@{}c@{}} 
                \textbf{Successful } \\
                \textbf{runs$^\dagger$}
            \end{tabular} & 
            \begin{tabular}[c]{@{}c@{}} 
                \textbf{Mean test set} \\
                \textbf{satisfaction (\%)}
            \end{tabular} \\
            \midrule
                Static input regression & 7/10 & 100 \\ 
            \midrule
               Dynamic input regression & 4/10 & 94 \\
            \midrule
                Feedback control & 8/10 & 99 \\
            \midrule
        \end{tabular}%
        \begin{tablenotes}
        \footnotesize
            \item[\dagger] Runs reaching \ac{stl} satisfaction $\ge 90\%$ in the training set.
        \end{tablenotes}
    \end{threeparttable}
\end{table}

In the third example, we train a \ac{bnn} to operate in closed-loop with an immune system model adapted from \cite{barber2021-computationalmodelsepsis} to represent chronic inflammation (see Fig.~\ref{fig:control-results}). The model consists of four species:  bacteria $X_B$, pro-inflammatory proteins $X_P$, anti-inflammatory proteins $X_A$, and tissue damage marker $X_D$. It is parametrized by $p$, which represents the rate at which $X_P$ can fight $X_B$. In the chronic state, $x_P$ stays elevated, driving tissue damage $X_D$. Treatments that increase $x_A$ reduce $x_P$ (and thus $x_D$), but also weaken defense against $X_B$. To address this, we design a \ac{bnn} controller that suppresses $X_P$ only when no $X_B$ is present, and allows a temporary rise in $x_P$ during infection to ensure $X_B$ never persists for more than 15h. This is formalized as:  
$\varphi_3: (
\event_{[0, \infty]} (\always_{[0, \infty]} X_D<150))\land (\lnot \event_{[0, \infty]} (\always_{[0, 15]}X_B>0.1))$.

We implement a one-layer \ac{bnn} with a single perceptron that takes $x_B$ and $x_D$ as inputs, and produces an action $u$ that increases $x_A$. The training set is generated by linearly sampling $20$ values of $p\in [12,15]$, and $20$ values of the initial concentration of bacteria $x_{B0}\in [0, 500]$. The test set is generated by shifting the grid 0.5 units for $p$, and 25 units for $X_{B0}$. Trajectories are simulated for $t=[0, 50, 100, \ldots, 800]$, with infection introduced at $t=200$ ($X_B(t=200)=X_{B0}$). 

We perform $10$ optimization runs with different random seeds, each for a maximum of $200$ iterations with a learning rate of $5\cdot10^{-3}$. In $8$ runs, $\varphi$ is satisfied by at least $90\%$ of the training trajectories, with an average satisfaction of $\varphi$ of $99\%$ in the test set. Fig.~\ref{fig:control-results} compares the behavior of the system under (1) the optimized \ac{bnn} controller, (2) a bacteria-unaware, which is the optimized \ac{bnn} but with its input for $x_B$ fixed to zero, and (3) no controller. In the untreated case, tissue damage accumulates due to persistent inflammation. The bacteria-unaware controller reduces damage but fails to clear the infection. The optimized controller allows temporary inflammation to eliminate bacteria, and then suppresses it to protect tissue.

\section{Discussion and Conclusion} \label{sec:discussion}

We propose a framework to optimize \acp{bnn} from \ac{stl} specifications for regression and feedback control tasks. This enables training without numerical target data, which is often unavailable in molecular biology. The algorithm uses gradient descent for efficiency and scalability, and we demonstrate its effectiveness on three biological case studies.

One limitation of our approach is that the evaluation of the optimized \ac{bnn} is purely empirical: \ac{stl} satisfaction is measured on the unseen test set, following common machine learning practice. While this provides a measure of generalization, it does not provide theoretical guarantees, which remains an important direction for future work. Further, the model of \acp{bnn} is theoretical, so \acp{bnn} may not behave experimentally as predicted by the model. Since high-fidelity computational models of \acp{bnn} are not yet established in the literature, we plan to improve robustness to biological variability by incorporating stochasticity and noise into the model, and by extending the \ac{stl} specifications to probabilistic formulations \cite{sadigh2015-safecontroluncertainty}. Additionally, the \ac{bnn} architectures used are densely connected. In practice, minimal connectivity and size are preferable for implementation. We plan to extend the optimization to promote sparsity and smaller networks.  Finally, the optimization is sensitive to initialization, as shown in Tab.~\ref{tab:control-results-table}.  We hypothesize that the non-smooth robustness formulations and discontinuities in the loss function may limit the convergence. In future work, we will explore alternative loss functions based on smooth \ac{stl} formulations to improve convergence.

\section{Acknowledgments}
We thank Jean Disset, Charles van de Mark, George Wachter, Jonathan Babb, and Jessica Louie for insightful discussions. This work was funded by the AFOSR under grant FA5590-23-1-0529, and by the NSF under grants NSF EFRI BEGIN OI 2422282, and NSF GCR 2219101.

\bibliographystyle{IEEEtran}
\bibliography{references}

\end{document}